\documentclass[11pt]{article}

\usepackage[final]{acl}

\usepackage{times}
\usepackage{latexsym}

\usepackage[T1]{fontenc}
\usepackage[utf8]{inputenc}

\usepackage{microtype}

\usepackage{inconsolata}

\usepackage{multirow}
\usepackage{booktabs}
\usepackage{graphicx}
\usepackage{amsmath}
\usepackage{amsfonts}
\usepackage{makecell}
\usepackage{hyperref}
\usepackage{stfloats}
\usepackage{tablefootnote}
\usepackage{algorithm}
\usepackage{algorithmic}

\usepackage{color}
\usepackage{soul}
\usepackage{makecell}
\usepackage{multicol}
\usepackage{supertabular}
\usepackage{listings}
\usepackage{arydshln}
\usepackage{tocloft}

\definecolor{lightblue}{rgb}{.8,.8,1}

\title{Self-Evolving GPT: A Lifelong Autonomous Experiential Learner}

 \author{Jinglong Gao$^1$\quad
Xiao Ding$^1$\footnotemark[1]\quad
Yiming Cui$^2$\quad
Jianbai Zhao$^1$\\
{\bf Hepeng Wang}$^1$ \quad
{\bf Ting Liu}$^1$ \quad
{\bf Bing Qin}$^1$\\
$^1$\normalsize{Research Center for Social Computing and Information Retrieval}\\[-.05cm]
\normalsize{Harbin Institute of Technology, China}\\[-.05cm]
$^2$\normalsize{State Key Laboratory of Cognitive Intelligence}\\[-.05cm]
\normalsize{iFLYTEK Research, Beijing, China}\\[-.05cm]
{\small\tt\{jlgao, xding, jianbaizhao, hpwang, tliu, qinb\}@ir.hit.edu.cn}\\[-.05cm]
{\small\tt ymcui@iflytek.com}}

\begin{document}
\maketitle
\begin{abstract}

\renewcommand{\thefootnote}{\fnsymbol{footnote}}
\footnotetext[1]{Corresponding Author}

To improve the performance of large language models (LLMs), researchers have explored providing LLMs with textual task-solving experience via prompts. However, they rely on manual efforts to acquire and apply such experience for each task, which is not feasible for the growing demand for LLMs and the variety of user questions.
To address this issue, we design a lifelong autonomous experiential learning framework based on LLMs to explore whether LLMs can imitate human ability for learning and utilizing experience. It autonomously learns and accumulates experience through experience transfer and induction, categorizing the types of input questions to select which accumulated experience to employ for them.
Experimental results on six widely used NLP datasets show that our framework performs reliably in each intermediate step and effectively improves the performance of GPT-3.5 and GPT-4. This validates the feasibility of using LLMs to mimic human experiential learning and application capabilities. Additionally, we provide a detailed analysis of the behavior of our framework at each step.

\end{abstract}

\section{Introduction}
Recently, large language models (LLMs) like ChatGPT have achieved excellent performance in various NLP tasks \citep{Koco__2023,ye2023comprehensive}.
However, numerous NLP tasks still cannot be effectively addressed by them \citep{mao2023gpteval,chang2023survey}. This is mainly because they have not accumulated enough experience to handle these tasks during their training.

\begin{figure}[t]
  \centering
\includegraphics[width=1\linewidth,, trim={1.4cm 1.9cm 1.4cm 0cm}, clip]{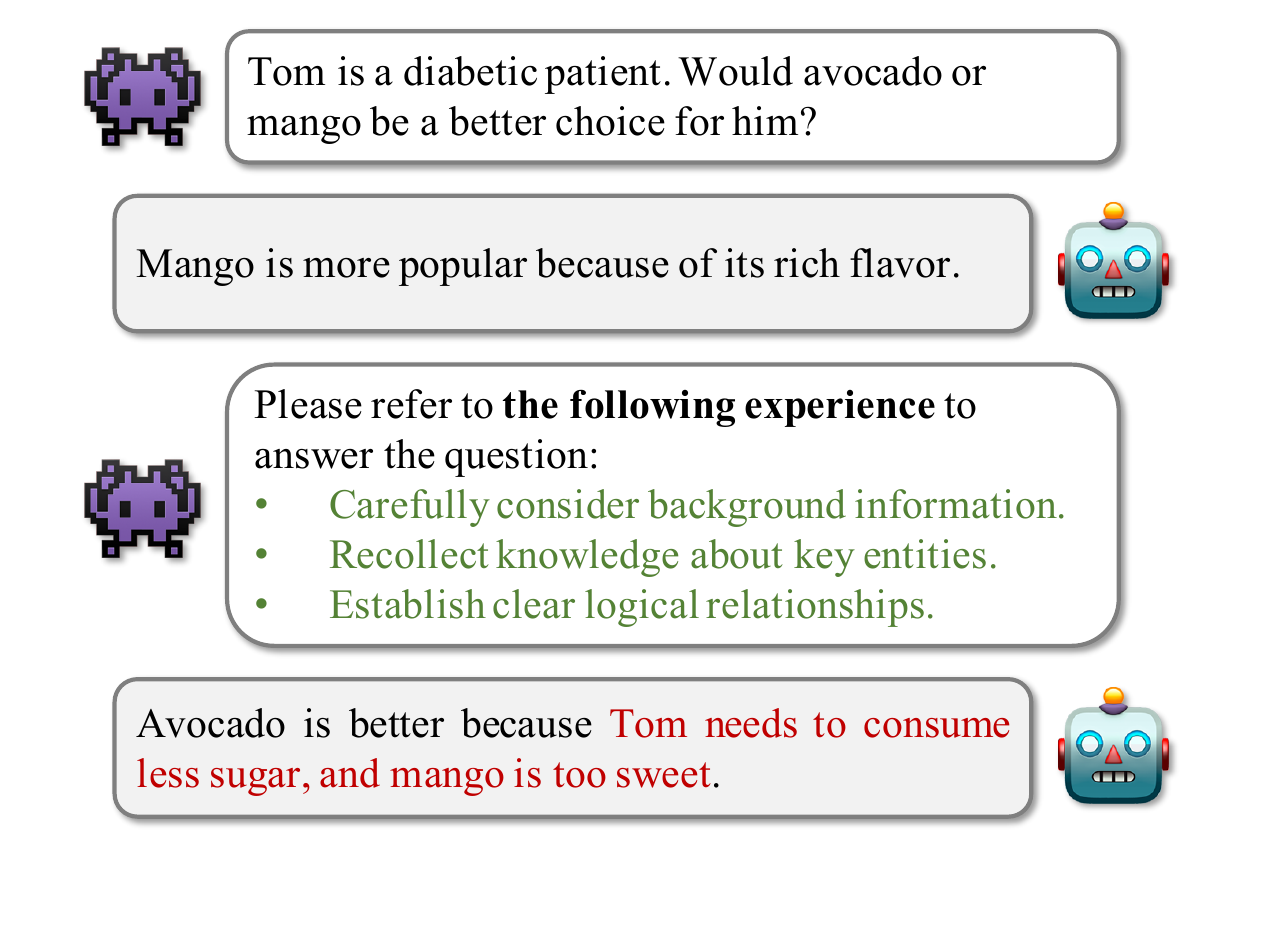}
  \caption{An example of experience-enhanced LLMs inference.}
  \label{fig:intro}
\end{figure}

To address these issues, previous studies have explored injecting task-solving experience into LLMs during the inference stage via prompts (as shown in Figure~\ref{fig:intro}). Their experience is textual descriptions of the task-solving processes, guidelines, and other insights.
Some studies manually craft such experience \citep{wei2022chain,kong2023better}.
Others attempt to summarize experience from manually annotated task datasets \citep{chen2023introspective,zhao2023expel,chen2024grimoire}, and then during inference, they essentially need to manually select the experience to apply to each question.
However, the demands of users on LLMs are ever-expanding, and the types of user questions continue to grow.
These methods would lead to high and unbounded costs for human labor.

In contrast, humans are capable of autonomous learning and utilizing experience.
Humans categorize encountered problems into different task types and induce experience from multiple concrete task practices, which are reused when encountering new problems of the same task type \citep{novak1984learning,cox1996introspective}. Besides, humans can transfer experience between similar tasks, thus gaining more experience without time-consuming practices \citep{deese1952psychology,perkins1992transfer}. As lifelong autonomous experience accumulates, humans gradually achieve ability growth.
Inspired by this, we want to explore whether LLMs can mimic the above process.
This could avoid the substantial manual labor and provide a unique evolutionary path for artificial general intelligence.

To facilitate this, we propose a lifelong autonomous experiential learning framework called Self-Evolving GPT (SE-GPT), which consists of a task-specific experience memory and five experience-centric modules based on ChatGPT.
For any user question, SE-GPT automatically categorizes the target task type and responds to the question with the target task experience in the memory.
For newly encountered task types, it learns experience through experience transfer and induction before responding.
Firstly, it locates similar tasks in its memory and transfers their experience to the target task. Then, it autonomously references web information and the transferred experience to practice the target task multiple times, thereby inducing more experience from its successes and failures.
Finally, the transferred and induced experience is added to the memory.
For tasks encountered previously, it assesses the need for repeating experience transfer and induction before responding, taking into account its proficiency level with the task.

To conduct experiments, we provide a basic implementation of our framework. We mainly focus on the overall framework and aim to analyze its effectiveness and behavior.
Experiments show that our framework is practically feasible.
It effectively improves the average performance of GPT-3.5 and GPT-4 on six widely used datasets by 3.8\% and 5.3\%, respectively.
Our framework reliably executes each intermediate module, achieving consistent performance improvements. Besides, we provide a detailed analysis of the behavior of our framework in each intermediate step.

\section{Related Work}

\subsection{Autonomous Experiential Learning}
To improve the performance of LLMs, researchers provide textual experience to LLMs through prompts. Early studies primarily involve manually crafting such experiential prompts \citep{wei2022chain,kong2023better}, while more recent work focuses on utilizing the LLMs themselves to obtain task-solving experience automatically.

Some studies focus on how to guide LLMs to automatically summarize experience based on interactive environments.
\citet{chen2023introspective} guided LLMs to summarize cooking skills in a cooking simulation game.
\citet{wang2023voyager} and \citet{zhu2023ghost} built LLM-based frameworks in the game ``Minecraft'' to autonomously learn to complete various game targets.
\citet{Generative} created a sandbox environment similar to ``The Sims'' to guide LLMs in learning role-playing skills.
Both \citet{wen2023dilu} and \citet{fu2024drive} taught LLMs how to perform autonomous driving in a simulated driving environment.

All of these studies guide LLMs to learn experience based on explicit feedback from environments, which is inaccessible for most NLP tasks.
Besides, they require human labor to create the environment or develop feedback-reading methods.

For NLP tasks, \citet{zhao2023expel} and \citet{chen2024grimoire} leveraged ChatGPT to automatically summarize experience from manually annotated NLP datasets.
\citet{zhao2023expel} employed Reflexion \citep{shinn2023reflexion} to generate reasoning chains for each question. Then, the experience is summarized from the questions, chains, and human-annotated labels by ChatGPT.
They also found that ChatGPT could transfer the summarized experience from the HotpotQA \citep{yang2018hotpotqa} dataset to the FEVER \citep{thorne2018fever} dataset.
\citet{chen2024grimoire} analyzed the impact of different examples and prompts on the quality of the summarized experience.

However, these methods still require human labor to obtain experience and determine which experience to employ for the current question.
In contrast, our framework autonomously learns and selects experience, saving many human labor costs.

\subsection{Unsupervised In-Context Learning}

In-Context Learning (ICL) provides demonstrations to LLMs, which can be regarded as a specific substitute for textual experience.
Therefore, we introduce the recent work on unsupervised ICL.

Several studies aim at predicting labels with LLMs for unlabeled questions, yielding demonstrations \citep{li2023mot,wan-etal-2023-better,zhang2023automatic}.
However, these studies still necessitate manual effort for the generation of questions.
Therefore, \citet{lyu-etal-2023-z} directly leveraged retrieved web texts as unlabeled questions, which is only suitable for specific task datasets.
In contrast, our framework is task-agnostic and designed to operate autonomously.

Furthermore, several studies employed LLMs to generate entire demonstrations \citep{Kim2022SelfGeneratedIL,yu2023thought,chen-etal-2023-self}.
SG-ICL \citep{Kim2022SelfGeneratedIL} requires the development set for selecting demonstrations, while TP-ICL \citep{yu2023thought} is designed explicitly for complex reasoning tasks like shortest-path reasoning, and Self-ICL \citep{chen-etal-2023-self} is the general-purpose one.
These demonstrations suffer from issues such as incorrect formatting, noise, and low diversity.
However, our framework utilizes the general insights summarized from multiple demonstrations, which is more reliable than the demonstrations themselves.

\section{Methodology}
\begin{figure*}[t]
  \centering
  \includegraphics[width=0.9\linewidth, trim={0 2cm 0 0.8cm},clip]{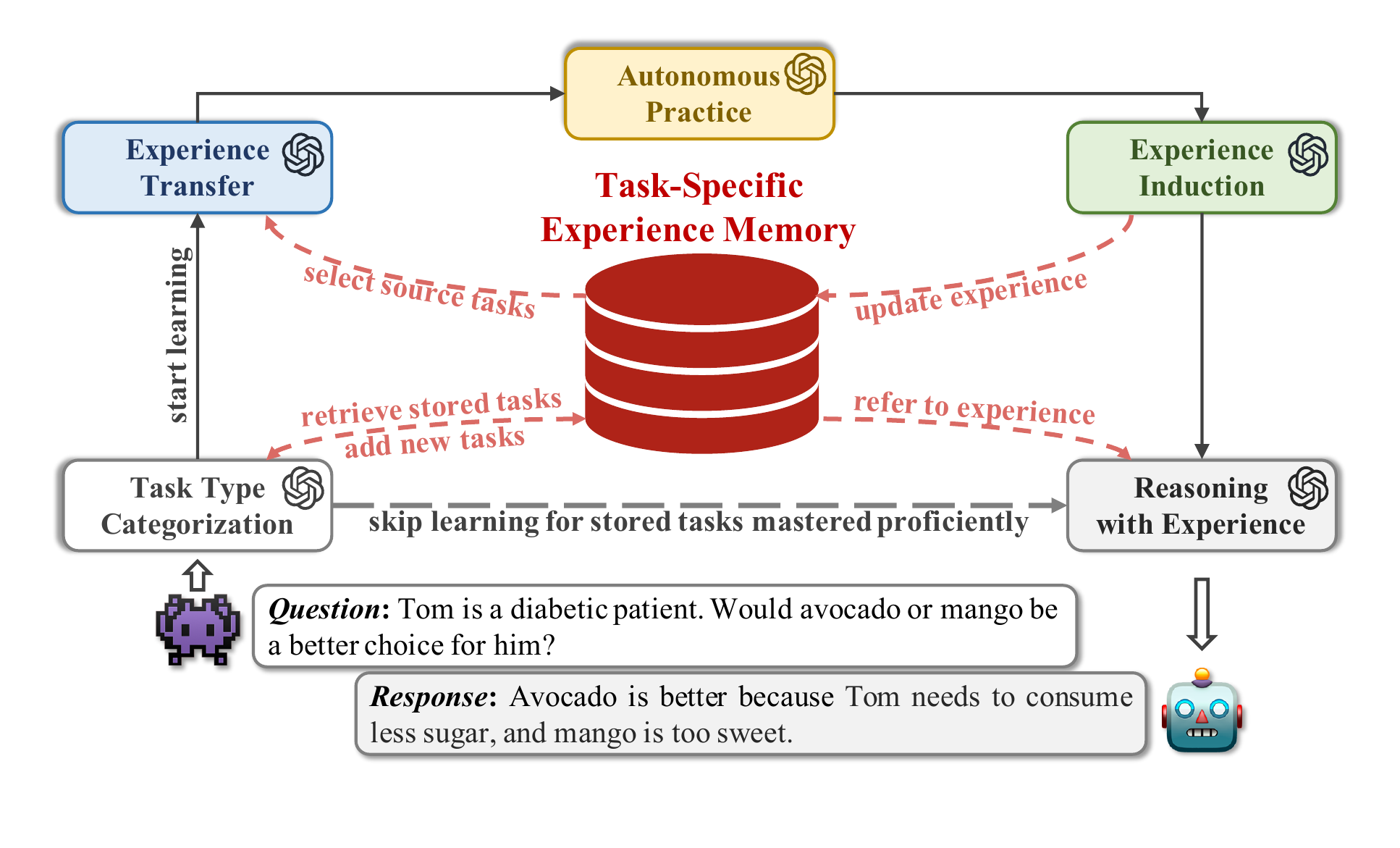}
  \caption{The framework of our proposed Self-Evolving GPT. The lines connected to the memory indicate the flow of information stored in memory. Other lines with arrows represent the execution sequence of our framework.}
  \label{fig:framework}
\end{figure*}

Figure~\ref{fig:framework} shows the framework of our proposed Self-Evolving GPT, which consists of one task-specific experience memory and five experience-centric modules based on ChatGPT.
Our framework continuously receives various user questions.
It automatically categorizes the task type of the question, and adds it to memory if it is a new task not yet stored.
For tasks that are not proficiently mastered, it performs experience transfer, autonomous practice, and experience induction to update their experience in memory.
Finally, it refers to experience stored in memory to respond the user question.

In practice, we provide a basic implementation of our framework, which may be further optimized. We primarily focus on the overall framework, and aim to analyze its effectiveness and behavior.
\textbf{The prompts and execution examples of our implementation are presented in Appendix~\ref{appendix:prompts} and~\ref{appendix:case_study_our}.}

\subsection{Task-Specific Experience Memory}
We utilize an external memory to store the task-specific textual experience that our framework autonomously learns.
This memory starts empty and gradually grows as our framework runs, assisting it in task-solving and learning new experience.

Specifically, we store each task in the memory with its name, description and experience. For the completeness of experience, our memory stores two types of experience for each task:
1) \textbf{Procedure}: the specific steps for handling the task;
2) \textbf{Suggestions}: how to better accomplish the task or avoid low-quality responses.
These task names, descriptions, and experience are all autonomously generated by~our framework.

\subsection{Task Type Categorization}
\label{sec:tti}

Users may pose various questions to the framework, corresponding to unpredictable task types. Therefore, we employ this module to first autonomously categorize the task type of each user question.

The operation of this module is divided into three steps:
1) ChatGPT utilizes Prompt~\ref{pt:p1} to generate the task name and description based on the question;
2) we retrieve the top 5 tasks from memory that are semantically most similar to the generated task description;
3) finally, ChatGPT utilizes Prompt~\ref{pt:p2} to select which one of the five tasks is identical to the generated task. If a match is found, the question is linked to the selected task; otherwise, it is linked to the generated task, and we add the generated task into the memory with empty initial task experience. Please note that the word ``task'' in our framework represents a ChatGPT-generated task rather than a classic NLP task (e.g., sentiment analysis) in a certain predefined task list.

After this, we retrieve the experience of the current task from memory, and denote it as $\mathbf{E}_\text{mem}$.
Then, we assess whether the current task has been adequately learned following our skip learning condition (\S\ref{skip_learning_condition}).
If it has, we respond to the user question with $\mathbf{E}_\text{mem}$ following our final reasoning prompt (\S\ref{sec:final_r}); otherwise, we learn experience following our experience transfer module (\S\ref{sec:transfer}), autonomous practice module \S\ref{sec:practice} and experience induction module \S\ref{sec:exp_induction}.

\subsection{Experience Transfer}
\label{sec:transfer}
Experience from similar tasks often exhibits transferability \citep{deese1952psychology,perkins1992transfer}.~Therefore, we employ this module to transfer the experience of other tasks in memory to the current task.

This module is orchestrated through four fundamental steps:
1) we retrieve the top 10 tasks from memory that are semantically most similar to the target task description;
2) if the previous step outputs at least one candidate task, ChatGPT utilizes Prompt~\ref{pt:p3} to select which among the 10 tasks should be chosen as source tasks for the transfer;
3) if the previous step outputs at least one source task, ChatGPT utilizes Prompt~\ref{pt:p4} to facilitate a step-by-step experience transfer process. It begins by understanding the differences between the source and target tasks, then identifying shared general experience between them, and finally rephrasing the general experience in the context of the target task. We denote such experience as $\mathbf{E}_\text{transferred}$;
4) if $\mathbf{E}_\text{mem}$ is not empty, ChatGPT utilizes Prompt~\ref{pt:p5} to merge $\mathbf{E}_\text{transferred}$ and $\mathbf{E}_\text{mem}$ for updating $\mathbf{E}_\text{transferred}$.
If steps 1 and 2 fail to select any source tasks, $\mathbf{E}_\text{mem}$ is employed as $\mathbf{E}_\text{transferred}$.

\subsection{Autonomous Practice}
\label{sec:practice}
Humans can autonomously practice tasks and derive experience from practice instances. Therefore, we employ this module to mimic the process of human autonomous practice. For the current target task, it automatically generates multiple examples, including questions, responses, and labels indicating whether the responses are correct.
Additionally, it utilizes the transferred experience and the autonomously retrieved web information to provide references for its practice process.

This module performs autonomous practice step by step:
1) we retrieve web documents that are semantically most related to the user question;
2) ChatGPT utilizes Prompt~\ref{pt:p6} to reference one of the retrieved web documents, the user question, and the task description generated in \S\ref{sec:tti} to generate a new question;
3) ChatGPT utilizes Prompt~\ref{pt:p7} to respond to the generated new question with $\mathbf{E}_\text{transferred}$;
4) ChatGPT utilizes Prompt~\ref{pt:p8} to reference the web document in the second step for verifying the correctness of its responses.
We repeat the above steps to obtain five examples for the current task.

\subsection{Experience Induction}
\label{sec:exp_induction}
After the autonomous practice, we summarize new experience for the current task from examples generated in \S\ref{sec:practice} with correct or incorrect answers.

In practice, we utilize Prompt~\ref{pt:p9} to guide ChatGPT in summarizing experience step-by-step. ChatGPT first summarizes the commonalities in the correct examples, identifying patterns in the incorrect examples, and compares the differences between the correct and incorrect examples.
Then, based on these observations and analysis, ChatGPT tries to summarize task-solving insights generally applicable to unseen examples of the current task. We denote such experience as $\mathbf{E}_\text{induced}$.
After that, if $\mathbf{E}_\text{transferred}$ is not empty, we utilize Prompt~\ref{pt:p5} to merge $\mathbf{E}_\text{induced}$ and $\mathbf{E}_\text{transferred}$ for updating $\mathbf{E}_\text{induced}$.

Finally, we replace $\mathbf{E}_\text{mem}$ in memory as $\mathbf{E}_\text{induced}$, which has been enhanced through experience transfer, autonomous practice and experience induction.

\subsection{Learning or Skip Learning}
\label{skip_learning_condition}
The tasks that our framework has already adequately learned do not require further learning.
It is inefficient to repeat learning for each user question.

Implementation-wise, our memory records the number of incorrect examples during each autonomous practice stage.
If the number of incorrect examples remains zero three times for the same task,
we consider that such task has already been adequately learned, and further learning is skipped.

Although we provide a basic skip condition, it may be modified for different preferences for efficiency and experience quality.

\subsection{Reasoning with Experience}
\label{sec:final_r}
Finally, we utilize Prompt~\ref{pt:p10} to guide ChatGPT in responding to the user question with the experience of the current task in memory.
For tasks that require further learning, the experience~stored~in~memory~has been enriched through experience transfer, autonomous practice, and experience induction.

\section{Experiments}
\subsection{Datasets and Evaluation Metrics}
\label{datasets}

We conduct experiments on the mixture of the following six widely used NLP datasets, including:
1) MMLU \citep{hendrycks2020measuring}, which is a massive multitask test consisting of multiple-choice questions from various branches of knowledge, covering 57 tasks;
2) e-CARE \citep{du-etal-2022-e}, which is a causal reasoning dataset that requires determining which option is the cause or result of a given event from various domains;
3) SocialIQA \citep{sap-etal-2019-social}, which is a social commonsense test that focuses on reasoning about people's actions and their social implications in various social situations;
4) WinoGrande \citep{sakaguchi2021winogrande}, which is a robust commonsense reasoning dataset formulated as a fill-in-a-blank task with binary options;
5) HELP \citep{yanaka-etal-2019-help}, which is a natural language inference dataset that focuses on logical inferences licensed by phrase replacements, so-called monotonicity reasoning;
6) LogiQA-2 \citep{liu2023logiqa}, which is sourced from expert-written questions for testing civil servants, covering multiple types of deductive reasoning.

We randomly select $K$ data points from each dataset and mix them randomly as the test dataset.
The test dataset includes human annotated labels, which are only used for evaluating performance.
For GPT-3.5, $K$=1,000, resulting in a final experimental data size of 6,000.
For GPT-4, $K$=500, resulting in a final experimental data size of 3,000. We adopt accuracy (Acc) as the evaluation metric and report the average accuracy of three rounds of predictions to reduce randomness.
For the human evaluation in our experiments, three evaluators are asked to perform annotations.

\subsection{Parameters Setting}
We conduct experiments using OpenAI's official API\footnote{\url{https://platform.openai.com/}} with two versions of ChatGPT separately, including \texttt{gpt-3.5-turbo-1106} (GPT-3.5) and \texttt{gpt-4-1106-preview} (GPT-4). Moreover, \texttt{temperature} is fixed as 1.
The retrieval operations in \S\ref{sec:tti}, \S\ref{sec:transfer} and \S\ref{sec:practice} are accomplished by the Faiss index \citep{faiss}.
For the stability of Prompt~\ref{pt:p2} and~\ref{pt:p8}, we run them multiple times until one option is output twice, and then we select this option as the final output.
The web texts in \S\ref{sec:practice} are retrieved from Wikipedia and truncated to 512 tokens. 
If Prompt~\ref{pt:p8} outputs ``inconclusive'' for a generated question-answer pair, we discard it.

\subsection{Baselines}
In our experiments, we employ the following baseline methods:
1) \textbf{Zero-shot}, we directly feed the input question into ChatGPT;
2) \textbf{Zero-shot-CoT}, we add ``Let's think step by step'' at the end of each input question and then feed it into ChatGPT;
3) \textbf{Self-EXP}, we first utilize Prompt~\ref{pt:p11} to instruct ChatGPT to directly generate \textbf{exp}erience for each input question. Then, just like our framework, we utilize Prompt~\ref{pt:p10} to guide ChatGPT in responding to each input question with the experience generated for it;
4) \textbf{Self-ICL}~\citep{chen-etal-2023-self}, which first prompts ChatGPT to generate new questions following the input question. Subsequently, ChatGPT predicts pseudo-labels for the new questions via zero-shot prompting. Finally, it performs ICL for the input question with the pseudo-question-label pairs as demonstrations;
5) \textbf{Self-ICL-CoT}~\citep{chen-etal-2023-self}, which is a Chain-of-Thought-based variation of \textbf{Self-ICL}. It adds ``Let's think step by step'' at the end of new questions and the input question before predicting them. We faithfully replicated the methods of \citet{chen-etal-2023-self} according to their origin paper;
6) \textbf{Modified Self-ICL}, from the test dataset, we retrieve the top 5 examples with the highest semantic similarity for each test example, to replace the generated input question in the self-ICL;
7) \textbf{AutoP-ICL}, employs demonstrations generated by our autonomous practice module (\S\ref{sec:practice}) to perform in-context learning. Specifically, the pairs (new question, reasoning process) deemed correct by our auto practice module are concatenated with the user query as the prompt for LLMs.

\subsection{Main Results}

\begin{table*}[!t]
\small
\centering
\setlength{\tabcolsep}{5.0pt}
\begin{tabular}{l|lccccccc}
\toprule
\textbf{Model} & \textbf{Method} & \textbf{MMLU} & \textbf{e-CARE} & \textbf{SocialIQA} & \textbf{WinoGrande} & \textbf{HELP} & \textbf{LogiQA-2} & \textit{\textbf{Average}} \\
\midrule
\multirow{8}{*}{\textbf{GPT-3.5}} 
 & \textbf{Zero-shot} & 0.670 & 0.813 & \underline{0.754} & \underline{0.679} & 0.502 & 0.516 & \underline{0.656}  \\
 & \textbf{Zero-shot-CoT} & 0.666 & 0.802 & 0.751 & 0.675 & 0.516 & \underline{0.522} & 0.655\\
 & \textbf{Self-EXP} & \underline{0.673} & 0.773 & 0.712 & 0.658 & 0.509 & 0.515 & 0.640 \\
 & \textbf{Self-ICL} & 0.621 & 0.728 & 0.693 & 0.604 & 0.494 & 0.349 & 0.582 \\
 & \textbf{Self-ICL-CoT} & 0.615 & 0.742 & 0.696 & 0.619 & 0.507 & 0.350 & 0.588 \\
&\textbf{Modified Self-ICL} & 0.655& \underline{0.814}	& 0.746	& 0.674	&\underline{0.534}	& 0.510&  0.656\\
&  \textbf{AutoP-ICL}  &0.652 &	0.799	&0.735	&0.650	&0.504	&0.422& 0.627 \\

 & \textbf{SE-GPT (Ours)} & \textbf{0.708} & \textbf{0.857} & \textbf{0.792} & \textbf{0.693} & \textbf{0.557} & \textbf{0.556} & \textbf{0.694}\\
\midrule
\multirow{6}{*}{\textbf{GPT-4}}& \textbf{Zero-shot} & 0.796 & 0.828 & 0.788 & 0.812 & 0.608 & \underline{0.706} & 0.756\\
 & \textbf{Zero-shot-CoT} & 0.822 & 0.830 & 0.805 & \underline{0.833} & 0.628 & 0.686 & 0.767\\
 & \textbf{Self-EXP} & \underline{0.834} & \underline{0.846} & \underline{0.808} & 0.828 & 0.646 & 0.698 & \underline{0.777} \\
 & \textbf{Self-ICL} & 0.732 & 0.808 & 0.740 & 0.795 & 0.649 & 0.651 & 0.729 \\
 & \textbf{Self-ICL-CoT} & 0.788 & 0.820 & 0.734 & 0.826 & \underline{0.655} & 0.607 & 0.738\\
 & \textbf{SE-GPT (Ours)} & \textbf{0.850} & \textbf{0.869} & \textbf{0.835} & \textbf{0.848} & \textbf{0.690} & \textbf{0.761} &\textbf{0.809} \\
\bottomrule
\end{tabular}
\caption{\label{tab:main}Experimental results (\%) on the mixture of six datasets. \textbf{Bold} and \underline{Underlined} numbers represent the 1st and the 2nd best performance of two versions of ChatGPT on each dataset.
``Average'' denotes the mean accuracy across different datasets for each method.}
\end{table*}

\begin{table*}[t]
\small
\centering
\setlength{\tabcolsep}{5.0pt}
\begin{tabular}{l|lccccccc}
\toprule
\textbf{Model} & \textbf{Method} & \textbf{MMLU} & \textbf{e-CARE} & \textbf{SocialIQA} & \textbf{WinoGrande} & \textbf{HELP} & \textbf{LogiQA-2} & \textit{\textbf{Average}} \\
\midrule
\multirow{3}{*}{\textbf{GPT-3.5}} 
 & \textbf{SE-GPT (Ours)} & \textbf{0.708} & \textbf{0.857} & \textbf{0.792} & \textbf{0.693} & \textbf{0.557} & \textbf{0.556} & \textbf{0.694}\\
 & \textbf{- w/o transfer}& 0.697 & 0.843 & 0.771 & \underline{0.689} & 0.535 & 0.541 & 0.679 \\
& \textbf{- w/o induction} & \underline{0.703} & \underline{0.851} & \underline{0.779} & 0.678 & \underline{0.542} & \underline{0.547} & \underline{0.683} \\
\midrule
\multirow{3}{*}{\textbf{GPT-4}}
 & \textbf{SE-GPT (Ours)} & \textbf{0.850} & \textbf{0.869} & \textbf{0.835} & \textbf{0.848} & \textbf{0.690} & \textbf{0.761} &\textbf{0.809} \\
 & \textbf{- w/o transfer}& 0.841 & 0.853 & \underline{0.827} & 0.838 & 0.673 & 0.744 & 0.796 \\
& \textbf{- w/o induction} & \underline{0.846} & \underline{0.859} & 0.819 & \underline{0.841} & \underline{0.683} & \underline{0.756} & \underline{0.801} \\

 \bottomrule
\end{tabular}
\caption{\label{tab:ablation}
Performance (\%) of our framework with/without experience transfer and induction.}
\end{table*}

\begin{table}[t]
\small
\centering
\setlength{\tabcolsep}{4.5pt}
\begin{tabular}{l|lcccc}
\toprule
\multirow{2}{*}{\textbf{Model}} & \multirow{2}{*}{\textbf{Method}}& \multirow{2}{*}{ \textbf{Acc}  }& \multicolumn{3}{c}{\textbf{Experience}}\\
\cmidrule(lr){4-6}
& & &\textbf{Sug.} & \textbf{Pro.} & \textbf{All}\\
\midrule
\multirow{3}{*}{\textbf{GPT-3.5}}
& \textbf{SE-GPT (Ours)} & 0.998 & 7.8 & 6.2 & 14.0 \\
& \textbf{- w/o transfer}& 0.999 & 5.0 & 4.6 & 9.5 \\
& \textbf{- w/o induction}& 1.000 & 7.0 & 5.8 & 12.7 \\

\midrule
\multirow{3}{*}{\textbf{GPT-4}}
& \textbf{SE-GPT (Ours)} & 0.998 & 11.5 & 10.4 & 21.9 \\
& \textbf{- w/o transfer}& 0.999 & 8.2 & 7.4 & 15.6 \\
& \textbf{- w/o induction}& 0.999 & 9.2 & 8.4 & 17.6 \\
 \bottomrule
\end{tabular}
\caption{\label{tab:ablation_exp}The statistics and human-evaluated accuracy~(\%) of experience of our framework with/without experience transfer and induction. We report the average number of insights for experience across all tasks in our memory at the end of the runtime. ``Sug.'' means the suggestions. ``Pro.'' means the procedure. ``All'' means both of them.}
\end{table}

Table~\ref{tab:main} shows the results on the mixture of six NLP datasets. We find that:

Firstly, our SE-GPT achieves consistently better performance than baseline methods and improves the average performance of zero-shot GPT-3.5 and GPT-4 by 3.8\% and 5.3\%, respectively.
This is because our framework can effectively learn task-solving experience and select appropriate experience for the input question.

Secondly, across all datasets, our framework shows the most significant gains over zero-shot GPT-3.5 and GPT-4 on the HELP dataset, with improvements of 5.5\% and 8.2\%, respectively.
The reason may be that zero-shot ChatGPT performs worst on HELP, and additional guidance is more helpful for questions that the ChatGPT itself is~not~good at.

Thirdly, the performance of Self-EXP is unstable.
This is due to the quality of the experience it generates is unreliable, with errors, irrelevant information, or insights that LLMs cannot follow.
We conduct a case study in Appendix~\ref{appendix:case_study}.
The powerful capabilities of GPT-4 alleviate this issue.
However, our approach summarizes experience by observing patterns across specific examples and transferring shared insights from multiple source tasks to the target task. This allows our framework to learn highly task-relevant and more general experience.

Besides, the demonstrations generated by Self-ICL and Self-ICL-CoT cannot effectively enhance the performance of ChatGPT. There are mainly three reasons: 1) ChatGPT often generates new questions that are inconsistent with the format of the example;
2) there are errors in the reasoning chains and pseudo-labels predicted by ChatGPT;
3)~new questions directly generated by ChatGPT may be simple and lack diversity.
We conduct a case study on them in the Appendix~\ref{appendix:case_study}.
However, by referencing web texts, our SE-GPT improves the diversity of generated questions and verifies the correctness of responses. Additionally, we do not directly use specific examples for inference but extract general patterns from them, reducing the impact of noise.

Additionally, our framework outperforms the Modified Self-ICL. This is because we do not directly use specific demonstrations but summarize task-solving insights from them, reducing the impact of noise and providing more direct guidance.

Moreover, according to the results of AutoP-ICL, the performance gains of our framework is not largely due to web retrieval. In our framework, web texts are only utilized in the auto practice module. Web retrieval aids in checking the correctness of practice and provides necessary guiding signals for lifelong learning, but these signals cannot be directly applied to solving user queries. Our experience induction module further summarizes task-solving experiences from multiple practices, while the experience transfer module enables these experiences to assist with other similar tasks.

Furthermore, baseline methods need to generate demonstrations or experience for each question. However, our SE-GPT reuses the learned experience across different questions, resembling human thought processes.

\subsection{Effect of the Experience Transfer and Induction}

As shown in Table~\ref{tab:ablation} and Table~\ref{tab:ablation_exp}, we analyze the variations of our framework with/without the experience transfer and the experience induction module: 1) ``- w/o transfer'', directly skips the experience transfer module of our framework; 2) ``- w/o induction'', skips the experience induction module after $1/3$ of all test data in our experiments, i.e., 2,000 for GPT-3.5 and 1,000 for GPT-4.
Please note that our framework learns from the test data (only their inputs and not their labels) as it proceeds to the next instance.
In human evaluation, we randomly select the experience of 100 tasks from memory and then identify insights that are incorrect, unrelated to the tasks, or cannot be followed by LLMs to report the ``Acc''.
We find that:

Firstly, both experience transfer and induction contribute to the performance and the experience quantity of the overall framework. This is mainly because they can acquire experience for the target task by transferring from other tasks or summarizing from multiple examples, respectively.

Secondly, ``- w/o induction'' maintains an acceptable level of performance. This indicates that after running for some time, our framework can still achieve consistent improvement only through experience transfer, which is more cost-effective than experience induction.

Besides, our framework can generate high-quality experience. This arises from the fact that our framework references web texts to generate low-noise examples for summarizing experience, and leverage shared insights from multiple source tasks to obtain more reliable experience.

\subsection{Analysis of the Task Type Categorization}

\begin{table*}[t]
\small
\centering
\begin{tabular}{l|lcccccc}
\toprule
\textbf{Model} & \textbf{Type} & \textbf{MMLU} & \textbf{e-CARE} & \textbf{SocialIQA} & \textbf{WinoGrande} & \textbf{HELP} & \textbf{LogiQA-2}\\
\midrule
\multirow{3}{*}{\textbf{GPT-3.5}}
& \textbf{Generated Task} & 0.99 & 0.98 & 0.98 & 1.00 & 1.00 & 0.99 \\
& \textbf{Matched Task} & 0.94 & 0.92 & 0.96 & 0.97 & 1.00 & 0.94 \\
& \textbf{All Task} & 0.97 & 0.93 & 0.96 & 0.98 & 1.00 & 0.96 \\

\midrule
\multirow{3}{*}{\textbf{GPT-4}}
& \textbf{Generated Task} & 0.99 & 1.00 & 1.00 & 1.00 & 1.00 & 1.00 \\
& \textbf{Matched Task} & 1.00 & 1.00 & 0.99 & 1.00 & 1.00 & 1.00 \\
& \textbf{All Task} & 0.99 & 1.00 & 0.99 & 1.00 & 1.00 & 1.00 \\
\bottomrule
\end{tabular}
\caption{\label{tab:tti_acc}Human evaluation (\%) of the task type categorization module.}
\end{table*}

\begin{table*}[!t]
\small
\centering
\begin{tabular}{l|lcccccc}
\toprule
\textbf{Model}  & \textbf{MMLU} & \textbf{e-CARE} & \textbf{SocialIQA} & \textbf{WinoGrande} & \textbf{HELP} & \textbf{LogiQA-2}\\
\midrule
\textbf{GPT-3.5} & 0.719 & 0.960 & 0.969 & 0.995 & 1.000 & 0.846 \\
\textbf{GPT-4}  & 0.968 & 0.998 & 1.000 & 0.990 & 0.978 & 0.982 \\
\bottomrule
\end{tabular}
\caption{\label{tab:src_select}Human evaluation (\%) of source task selection.}
\end{table*}

\paragraph{Human Evaluation of Categorizing Task Types.}
Task type categorization is the first module of our framework and critically influences the performance of subsequent modules.
Table~\ref{tab:tti_acc} shows the human-evaluated accuracy of our task type categorization module. 
For each dataset, we randomly evaluate 100 questions linked to newly generated tasks and 100 questions matched to tasks in memory.
Accuracy on all data is reported as the weighted accuracy average for both. We find that ChatGPT performs very well in this stage. This is mainly due to it is not a difficult task, and we provide a reasonable prompt for ChatGPT.

\begin{figure}[t]
  \centering
\includegraphics[width=1\linewidth,, trim={0cm 0cm 0cm 0cm}, clip]{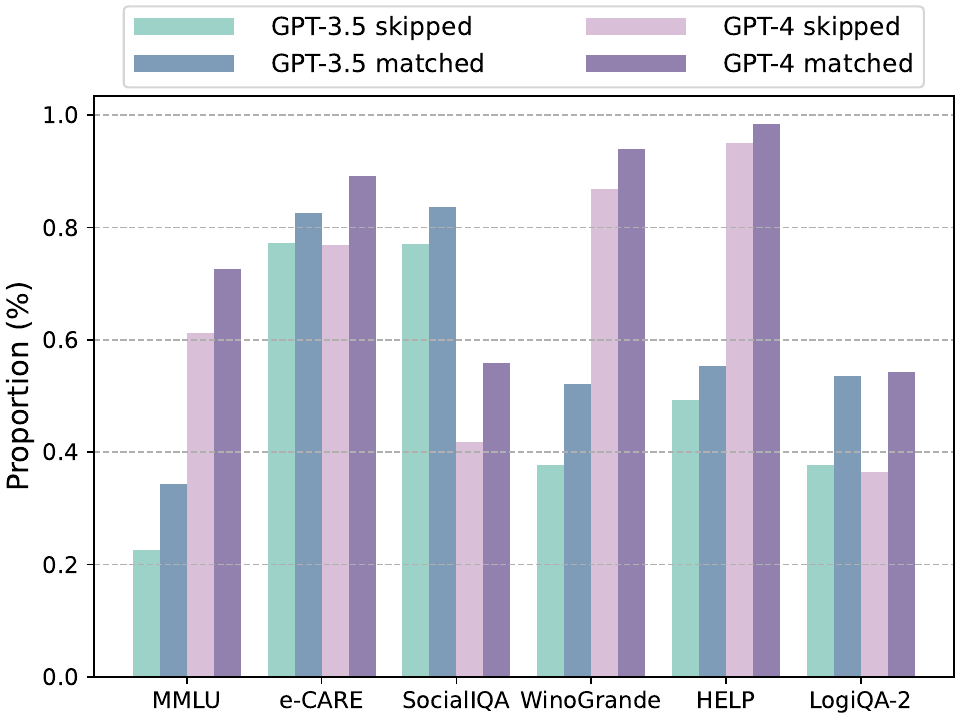}
  \caption{The proportion (\%) of the questions that match existing tasks in memory or skip the learning process.}
  \label{fig:tti_stat}
\end{figure}

\paragraph{Proportion of Matched and Skipped Questions.}\label{par:tti_prop}
Figure~\ref{fig:tti_stat} shows the proportion of the input questions that are matched to tasks in memory or skip the learning process. These proportions determine the efficiency of our framework in utilizing stored experience without the need to repeat the experiential learning process for each question.
We find that: 1) compared to GPT-3.5, more questions are matched and skipped by GPT-4. The main reason is the stronger capabilities of GPT-4, allowing it to better recognize learned tasks and meet the skipping criteria in \S\ref{skip_learning_condition}; 2) the trends in SocialIQA are opposite to those in other datasets. This may arise from the differences of ChatGPT in the prior knowledge and biases of task categorizing.

\subsection{Analysis of the Experience Transfer}
\paragraph{Human Evaluation of Selecting Source Tasks.} 
Table~\ref{tab:src_select} shows the human-evaluated accuracy of our source task selection process.
For each dataset, we randomly evaluate 100 target tasks, leading to 2,825 source-target task pairs.
We find that: 1) overall, ChatGPT performs well in selecting source tasks. This is mainly because recognizing similarity is not a difficult task; 2) the accuracy on MMLU is relatively low. This might arise from the diverse types of tasks in MMLU and its low similarity with other datasets. However, our framework still achieves improvements on MMLU. This is due to we identify shared insights among multiple source tasks, excluding non-transferable insights.

\paragraph{Number of Source Tasks Varying with Runtime.}\label{par:src_task_gpt3.5}

\begin{figure}[t]
  \centering
\includegraphics[width=1\linewidth,, trim={0cm 0cm 0cm 0cm}, clip]{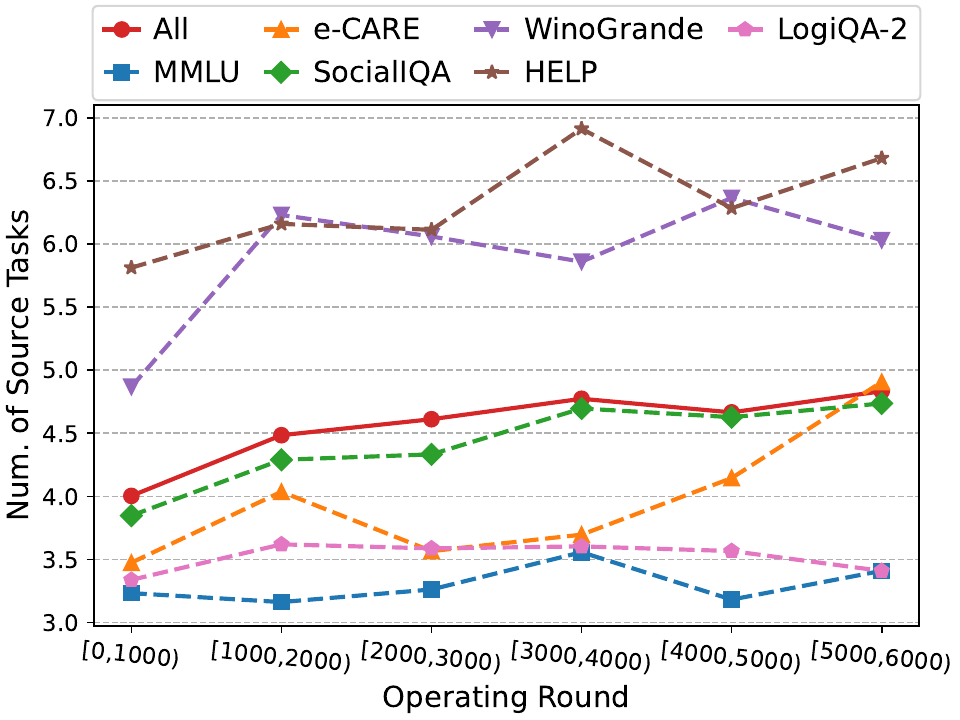}

  \caption{The average number of source tasks chosen per target task for experience transfer in each dataset during the execution of our SE-GPT based on GPT-3.5.}
  \label{fig:src_over_time}
\end{figure}

Figure~\ref{fig:src_over_time} shows the average number of source tasks of each input task varying with runtime.
The operating round refer to the number of test questions processed by our framework.
As the operating rounds increase, our framework can utilize more source tasks. The main reason is the increasing types of tasks in memory.
This also implies that our framework could continually enhance its transfer ability, benefiting from lifelong learning.

\begin{table}[t]
\small
\centering
\setlength{\tabcolsep}{6pt}
\begin{tabular}{lcccc}
\toprule
\multirow{2}{*}{\textbf{Method}} & \multirow{2}{*}{\textbf{Verify}}& \multicolumn{3}{c}{\textbf{Generated Question}}\\
\cmidrule(lr){3-5}
& &\textbf{Num} & \textbf{Dist-1}& \textbf{Dist-2}\\
\midrule
\multirow{2}{*}{\textbf{Ours}}& \multirow{2}{*}{0.877}& 5 & 0.31 & 0.51 \\
&& 20 & 0.12 & 0.25 \\
\multirow{2}{*}{\textbf{- w/o reference}}& \multirow{2}{*}{0.797
}& 5 & 0.24 & 0.41 \\
&& 20 & 0.03 & 0.08  \\
 \bottomrule
\end{tabular}
\caption{\label{tab:auto_practice}Performance of the autonomous practice module with GPT-3.5. ``Verify'' shows the human-evaluated accuracy (\%) of the validation step (Prompt~\ref{pt:p8}). ``Num'' is the count of questions generated per input question. ``Dist-n'' is the ratio of distinct N-grams to total N-grams in the generated questions per input question.}
\end{table}

\subsection{Analysis of the Autonomous Practice}

As shown in Table~\ref{tab:auto_practice}, we analyze the performance of the autonomous practice module with/without reference web texts. We randomly selected 300 generated examples and manually evaluate whether the validation results are correct. Besides, we report the diversity of new questions generated per input question. We find that by referencing web texts, our framework significantly improves both the validation accuracy and the diversity of generated questions. 
This is because: 1) the differences in reference texts lead to variations in generating questions; 2) the texts referenced by question generation usually contain question-solving information.

\subsection{Analysis of the Experience Induction}

\begin{figure}[t]
  \centering
\includegraphics[width=1\linewidth,, trim={0cm 0cm 0cm 0cm}, clip]{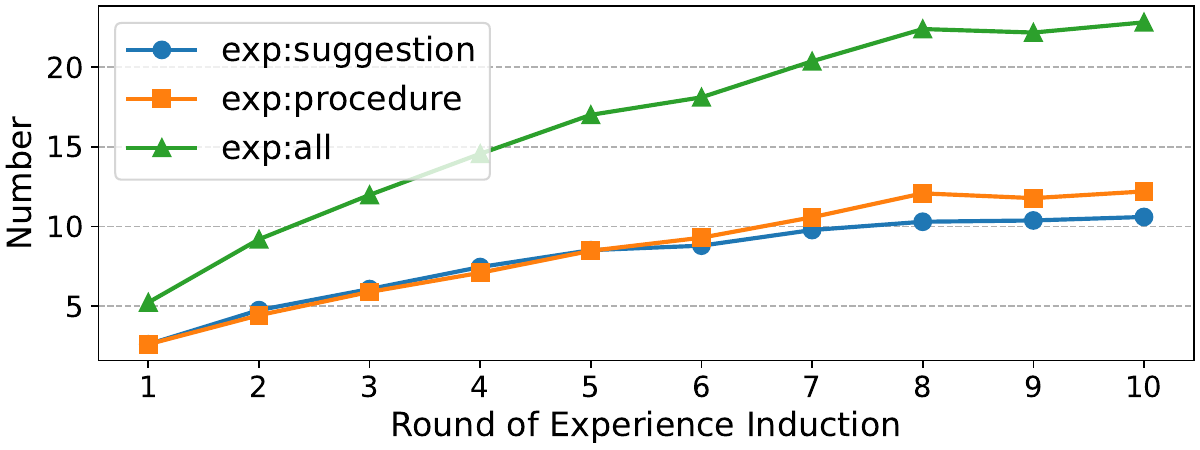}

  \caption{The number of insights generated by multi-round experience induction.}
  \label{fig:exp_induction_multi_turn}
\end{figure}

As shown in Figure~\ref{fig:exp_induction_multi_turn}, we repeatedly perform the autonomous practice and the experience induction module, reporting the number of generated insights.
We randomly select 100 questions and employ GPT-3.5 for the test. We find that the experience increases with each round and stabilizes at the 8th round.
This is because as the quantity of experience increases, the difficulty of acquiring new experience grows.
Through case observation, we find that almost all of the insights obtained in round 9 are included in the experience obtained previously.

\section{Conclusion \& Future Work}
In this paper, we propose a lifelong autonomous experiential learning framework based on LLMs.
It continuously and autonomously accumulates experience in solving tasks through experience transfer and induction, recognizing the nature of input questions to align them with relevant experience.
Considering the increasing demand for LLMs and the emergence of new types of user questions, our framework effectively reduces the human labor associated with previous methods.
Experiments show that the implementation of our framework can reliably execute each intermediate module and effectively enhance overall performance for responding to the input question. The following content may be subject to our research in future work: 1) \textbf{Enhanced engineering designs.} We only offer a basic implementation for our framework, and there is still room for improvement, e.g., supporting more complex functions; 2)~\textbf{Cold start.} At present, we run our framework completely from empty memory. However, the existing manually annotated datasets can be used to replace the autonomous practice module. Our framework can first learn from the manually annotated datasets, complete the cold start, and then run independently; 3) \textbf{Employing a combination of different-scaled LLMs to implement the framework.} It is evident that not all tasks necessitate using ChatGPT; integrating LLMs of various scales can achieve a balance between cost and performance; 4) \textbf{Experience Distillation.} Distilling the rich experience summarized by GPT-4 onto smaller-scale LLMs to enhance their performance on tasks that have been adequately learned by GPT-4.

\section*{Acknowledgments}

We would like to thank the anonymous reviewers for their constructive comments, and gratefully acknowledge the support of the National Natural Science Foundation of China (U22B2059, 62176079), and the Natural Science Foundation of Heilongjiang Province (YQ2022F005).

\section*{Limitations}
In this work, we design a framework to validate the feasibility of using LLMs to mimic human experiential learning and application capabilities. However, it is a basic implementation for experimental exploration but not a perfect LLM product, with room for improvement:
1) \textbf{Experience Failure and Operating Error:} Even with high-quality experience, LLMs may still make mistakes. Common errors we observed include reasoning errors/hallucination, LLMs disregarding partial experience, and LLMs lacking necessary knowledge to solve problems. Besides, the steps such as auto practice, experience induction and transfer are complex, and there still remains some noise in the obtained experience;
2) \textbf{Both Computationally and Financially Expensive:} the system repeatedly invokes an LLM, which is quite expensive both computationally and also financially. In \S\ref{promptcost}, we carefully discuss our prompt cost and possible methods to reduce the cost.
3) \textbf{Task Applicability:} Experience may still be effective in tasks requiring skills such as mathematical reasoning, but it might not be as effective for tasks relying on factual knowledge such as WikiQA. Therefore, the framework should have the ability to adaptively determine whether past experience is needed;

\bibliography{custom}

\clearpage

\appendix
\section*{Appendix}
\tableofcontents

\clearpage
\section{Additional Experimental Analysis}\label{appendix:addtion_analyze}
\subsection{Number of Source Tasks Varying with Runtime based on GPT-4}
\begin{figure}[b]
  \centering
\includegraphics[width=1\linewidth,, trim={0cm 0cm 0cm 0cm}, clip]{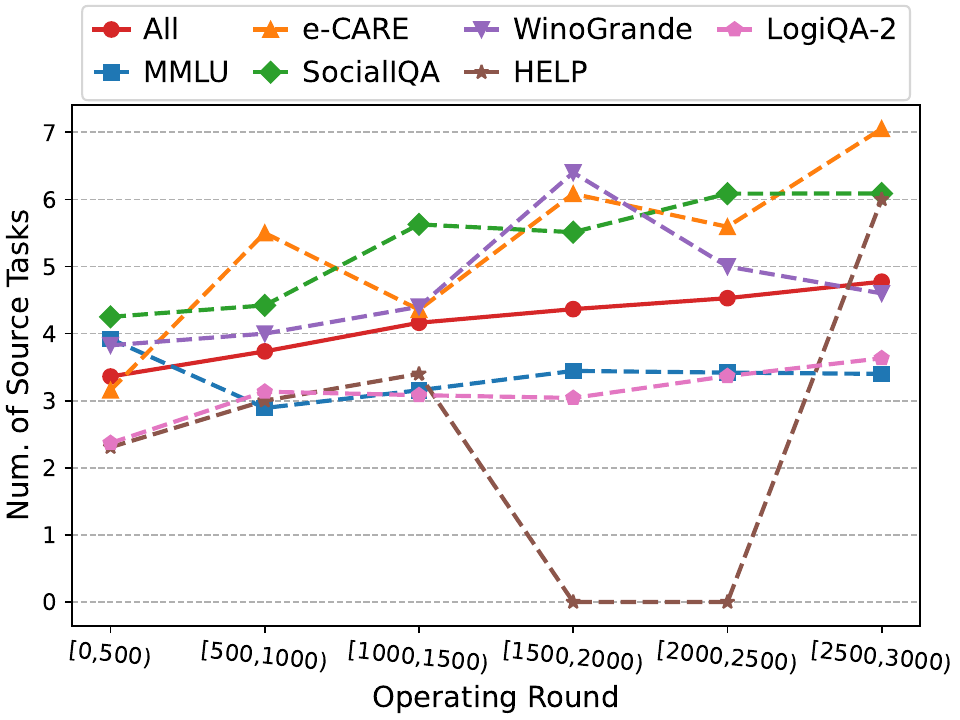}
  \caption{The average number of source tasks selected for each target task during the execution of our framework based on GPT-4.}
  \label{fig:source_task_gpt4}
\end{figure}

Figure~\ref{fig:source_task_gpt4} shows the average number of source tasks selected for each target task during the execution of our framework based on GPT-4. Overall, the performance of GPT-4 is consistent with the performance of GPT-3.5 that we analyzed in \S\ref{par:src_task_gpt3.5}.
An exception occurs with the HELP dataset, where the number of source tasks runs to 0 between 1500 to 2500 iterations. This is due to we do not consider input questions that skip learning when calculating the average number of source tasks. In other words, between 1500 to 2500 iterations, no examples in the HELP dataset require experience transfer. This is because the proportion of questions skipping learning is relatively high in the HELP dataset, as described in \S\ref{par:tti_prop}.

\subsection{Number of Tasks and Experience in the Memory Varying with Runtime}
\begin{figure}[t]
  \centering
\includegraphics[width=1\linewidth,, trim={0cm 0cm 0cm 0cm}, clip]{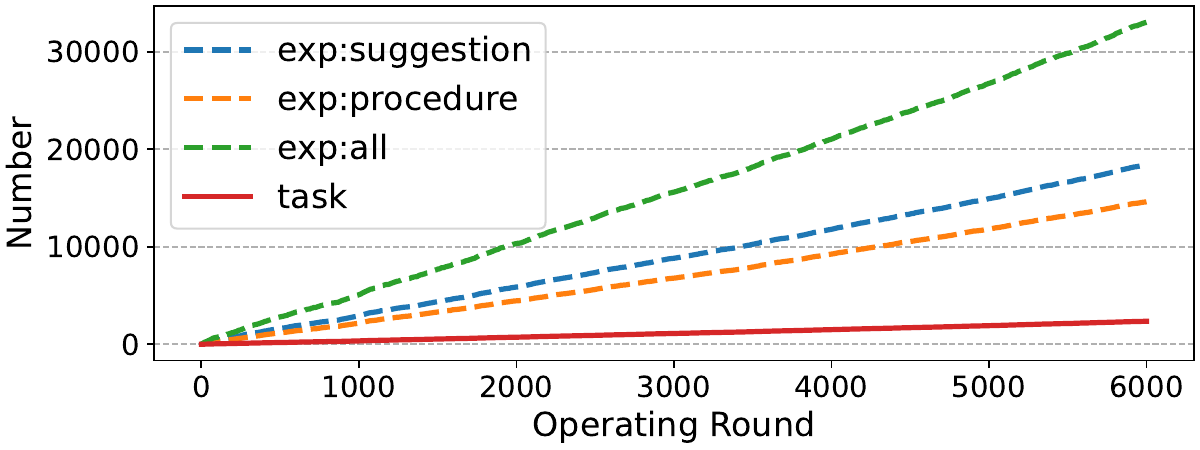}
  \caption{The number of insights and tasks in memory during the execution of our framework based on GPT-3.5.}
  \label{fig:te_time_gpt3.5}
\end{figure}

\begin{figure}[t]
  \centering
\includegraphics[width=1\linewidth,, trim={0cm 0cm 0cm 0cm}, clip]{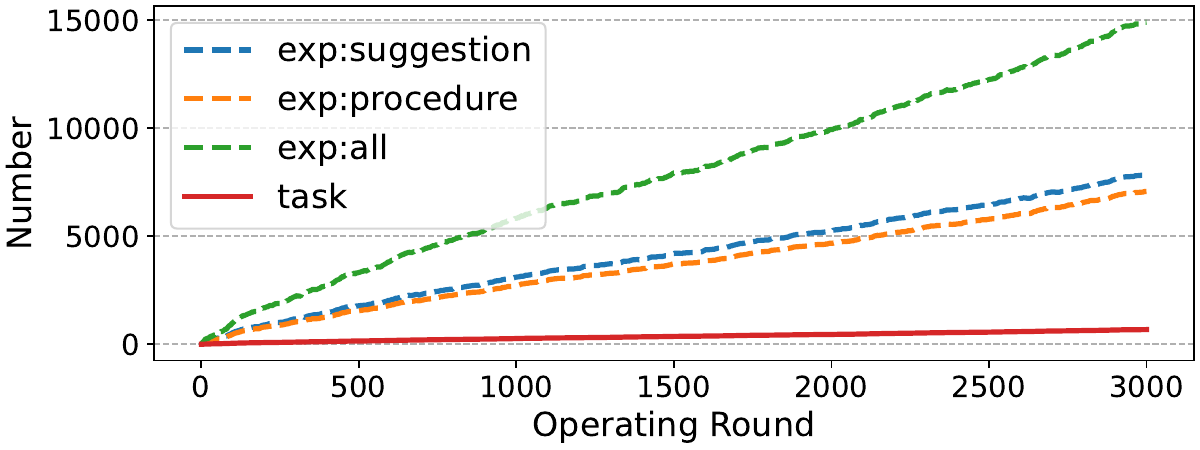}
  \caption{The number of insights and tasks in memory during the execution of our framework based on GPT-4.}
  \label{fig:te_time_gpt4}
\end{figure}

Figure~\ref{fig:te_time_gpt3.5} and Figure~\ref{fig:te_time_gpt4} show the number of insights and tasks in memory during the execution of our framework based on GPT-3.5 and GPT-4, respectively. We find that as the number of running rounds increases, our framework accumulates more task-specific experience. This indicates that the capabilities of our framework grow over time, enabling it to cover a broader range of user target tasks or provide experience for more user questions through experience transfer.

\subsection{Performance of Experience Induction Through More Examples}
\begin{figure}[b]
  \centering
\includegraphics[width=1\linewidth,, trim={0cm 0cm 0cm 0cm}, clip]{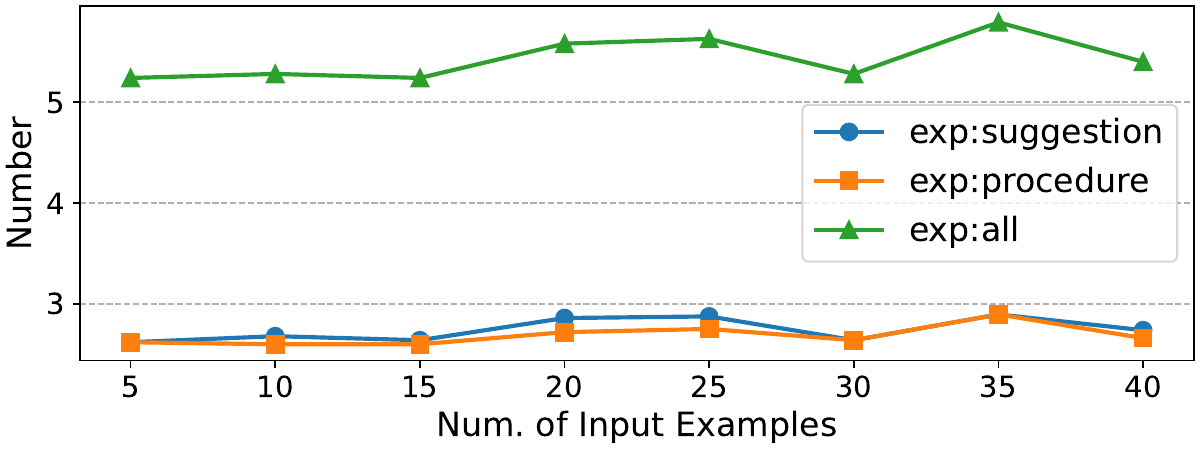}
  \caption{The number of insights generated by experience induction based on GPT-3.5 with more input examples.}
  \label{fig:exp_induction_more_dms}
\end{figure}

Figure~\ref{fig:exp_induction_more_dms} shows the number of experience generated by the experience induction module based on GPT-3.5 with more input examples.
It can be found that ChatGPT cannot effectively summarize more experience from a larger number of examples. This may be due to the increased difficulty for ChatGPT to analyze, requiring ChatGPT to think for a longer time.

\subsection{Prompt Cost}\label{promptcost}

As shown in Table~\ref{tab:prompt_cost}, we analyze the cost of our framework by reporting the average token usage per prompt for each example. Please note that for a single example, a prompt may be run multiple times due to reasons such as output format errors or API crashes. All these occurrences are included in the statistics to reflect the true cost.

It can be found that, compare to the traditional zero-shot CoT method, our framework are much more expensive and time-consuming. Overall, our main experiment using GPT-3.5 requires about five days of running, whereas GPT-4 requires three to five times longer. However, this does not mean that the framework is without hope for practical application, as our current basic implementation focuses more on demonstrating the behavior of LLMs at various stages, without any optimization for efficiency.
We believe the following approach can be further explored to reduce operational costs:
\begin{itemize}
    \item Use existing annotated corpora to replace the Autonomous Practice module. It can be found that the main cost of our framework lies in the Autonomous Practice module. Our previous experimental results indicate that, given sufficient prior experience, our framework can perform comparably to the complete framework solely through experience transfer. Therefore, allowing the framework to gain experience from existing annotated corpora first could significantly reduce the substantial costs associated with the Autonomous Practice module.
    \item Consider using smaller PLMs to perform simple steps. Within our framework, prompts 4, 9, and 10 are involved in experience transfer, experience induction, and experience application, respectively. Other steps are relatively simple and can be substituted with smaller PLMs instead of the expensive ChatGPT.
\end{itemize}

\begin{table*}[t]
\small
\centering
\setlength{\tabcolsep}{16pt}
\begin{tabular}{lcccc}
\toprule
\multirow{2}{*}{\textbf{Module}} & \multirow{2}{*}{\textbf{PROMPT}} & \multicolumn{3}{c}{\textbf{Usage}}\\
\cmidrule(lr){3-5}
& & \textbf{Input} & \textbf{Output}& \textbf{Total}\\
\midrule
\multirow{2}{*}{\textbf{Task Type Categorization}} & Prompt 1 & 260 & 63 & 323 \\
& Prompt 2 & 825 & 26 & 851 \\ \hdashline
\multirow{3}{*}{\textbf{Experience Transfer}} & Prompt 3 & 289 & 11 & 300 \\
& Prompt 4 & 2139 & 381 & 2520 \\
& Prompt 5 & 456 & 221 & 677 \\\hdashline
\multirow{3}{*}{\textbf{Autonomous Practice}} & Prompt 6 & 1209 & 386 & 1595 \\
& Prompt 7 & 1184 & 607 & 1791 \\
& Prompt 8 & 3292 & 309 & 3601 \\\hdashline
\textbf{Experience Induction}& Prompt 9 & 1145 & 183 & 1328 \\\hdashline
\textbf{Reasoning with Experience}& Prompt 10 & 532 & 15 & 546 \\\hdashline
\textbf{Total}& - & 11331 & 2202 & 13532 \\
\midrule 
\textbf{Zero-shot-CoT} & Zero-shot-CoT & 159 & 32 & 191 \\

 \bottomrule
\end{tabular}
\caption{\label{tab:prompt_cost} Average token usage per prompt for each example.}
\end{table*}

\clearpage

\section{Case Study}\label{appendix:case_study}
In this section, we analyze the case of Self-EXP and Self-ICL.

\subsection{Experience Generated by Self-EXP}
\label{appendix:case_self_exp}

\textbf{Self-EXP} employs Prompt~\ref{pt:p11} to instruct ChatGPT to directly generate experience for each input question.
Case~\ref{cs1} shows the example of experience output by Self-EXP based on GPT-3.5.

For the 1st example, we can find that although the generated experience seems reasonable, it is not well aligned with the input problem. In fact, these generated insights are wrong or irrelevant to the input problem.
The possible reason is that ChatGPT only focuses on the keywords of the input question without understanding the essential task objective and the processing flow of the task.

For the 2nd example, Self-EXP suggests that ``Consider Addison's typical preferences and behaviors'' and ``Ask Jesse about Addison's purpose.'' are valuable insights. These suggestions require LLMs to have the ability to actively explore unknown information and communicate with humans. However, ChatGPT itself does not possess such abilities, and implementing such abilities requires additional auxiliary modules.

Compared with Self-EXP, our framework generates multiple pairs of pseudo questions and reasoning processes, and summarizes experience from them.
This ensures that the experience generated by our framework is highly consistent with the input question and matches the abilities of LLMs.
Besides, this enables our framework to discover new insights rather than relying solely on the experience learned during ChatGPT's pre-training process.

\subsection{Demonstrations Generated by Self-ICL}
\label{appendix:case_self_icl}
\textbf{Self-ICL} prompts ChatGPT to generate demonstrations for each input question.
Case~\ref{cs2} shows the example of demonstrations generated by Self-ICL based on GPT-3.5. We can find that:

Firstly, there is a format inconsistency between the input question and the question generated by Self-ICL. In fact, this format issue is not an exception but often occurs in generated pseudo questions.
ICL will make such format errors interfere with the reasoning process of the LLMs.

Besides, the 2nd example contains a wrong pseudo-response.
This response error is difficult to avoid for LLMs unless they have fully mastered the task type of the input question.

In summary, the demonstrations generated by Self-ICL exhibit issues such as inconsistent formatting and noisy responses.
However, our framework does not directly utilize the generated demonstrations; instead, it extracts general experience from demonstrations, thereby mitigating the impact of the issues above.

\section{Additional Information on Responsible NLP Research}
\label{appendix:checklist}

\paragraph{Use Scientific Artifacts.} As shown in Section \S\ref{datasets}, we use six NLP benchmark datasets in English.
They are all allowed to be used for scientific research. They do not contain information that names or uniquely identifies individual people or offensive content.

\paragraph{Human Evaluators.}
We recruit NLP graduate students to conduct evaluation and are compensated by providing academic credits and guidance. They know explicitly that the annotated data is used for scientific research.

\paragraph{AI Assistants in Writing.}
We use ChatGPT to help check for grammatical errors and provide suggestions for improving language expression.

\captionsetup[table]{name=Case} 
\captionsetup{labelformat=empty} 
\setcounter{table}{0}

\begin{table*}[h] \small 
\centering
\begin{tabular}{p{\linewidth}}
\toprule
\textbf{Case 1: experience generated by Self-EXP.}\\
\midrule
\begin{lstlisting}[escapeinside={(*@}{@*)}]
(*@\textbf{[Example 1]}@*)

(*@\textbf{Input Question:}@*)
Sentence: I couldn't figure out what my costume would be for Halloween Peter Pan or Snow White. The _ costume I'd have to wear a dress.
Option A: Peter Pan
Option B: Snow White
Choose the more appropriate option to fill in the blank space in the given sentence.
Use the following JSON format to output your answer:
```json
{
  "correct option ID": /* one of A, B */
}
```

(*@\textbf{Suggestion:}@*)
- Consider the overall theme or context of the event you will be attending for Halloween.
- Think about which character you would enjoy portraying the most.
- Take into account any props or accessories you may need for the costume.
- Consider how comfortable you would be wearing a dress for the duration of the event.
- Avoid making a decision solely based on gender stereotypes.

(*@\textbf{Procedure:}@*)
- Research the character traits, appearance, and costumes of Peter Pan and Snow White.
- Consider your personal preferences and the level of comfort in each costume.
- Try on different costume options to see which one suits you best.
- Seek feedback from friends or family members.
- Ultimately, choose the option that aligns with your personal interests and comfort level.
\end{lstlisting}\\
\midrule
\begin{lstlisting}[escapeinside={(*@}{@*)}]
(*@\textbf{[Example 2]}@*)

(*@\textbf{Input Question:}@*)
Context: Jesse walked Addison's dog because she was on vacation.
Question: What will Addison want to do next?
Option A: great
Option B: going to the ocean to swim
Option C: loved
Based on the given context, choose the correct answer to the question from the three options.
Use the following JSON format to output your answer:
```json
{
  "correct option ID": /* one of A, B, C */
}
```

(*@\textbf{Suggestion:}@*)
- Understand the context of the situation and the characters involved
- Consider Addison's typical preferences and behaviors
- Ask Jesse about Addison's purpose.

(*@\textbf{Procedure:}@*)
- Analyze the situation - Addison's dog being walked by Jesse while she's on vacation - and anticipate what Addison would want next
- Choose the option that aligns best with Addison's typical behaviors and interests
\end{lstlisting}\\
\bottomrule
\end{tabular}
\caption{}
\label{cs1}
\end{table*}

\begin{table*}[h] \small % prompt
\centering
\begin{tabular}{p{\linewidth}}
\toprule
\textbf{Case 2: demonstrations generated by Self-ICL.}\\
\midrule
\begin{lstlisting}[escapeinside={(*@}{@*)}]
(*@\textbf{[Example 1]}@*)

(*@\textbf{Input Question:}@*)
Premise: The magazine revealed all his dirty watchwords .
Hypothesis: The magazine revealed all his dirty secrets .
You need to decide whether the hypothesis is entailed by the premise by choosing one of the following answers:
"Yes": The hypothesis follows logically from the information contained in the premise.
"No": The hypothesis is logically false from the information contained in the premise.
"Neutral": It is not possible to determine whether the hypothesis is true or false without further information.
Use the following JSON format to output your answer:
```json
{
  "answer": /* Yes, No or Neutral */
}
```

(*@\textbf{Generated Pseudo-Question:}@*)
Premise: The company announced a new product launch date.
Hypothesis: The new product will be a bestseller.
```json
{
  "answer": "Neutral"
}
```
(*@\textbf{Generated Pseudo-Response:}@*)
"Neutral" - The premise of the company announcing a new product launch date does not necessarily entail that the new product will be a bestseller. It is possible that the new product could be successful, but it is not guaranteed based on the premise alone.
\end{lstlisting}\\
\midrule
\begin{lstlisting}[escapeinside={(*@}{@*)}]
(*@\textbf{[Example 2]}@*)

(*@\textbf{Input Question:}@*)
Premise: Most of them chose to eat steak.
Choice A: A group of  men went to a restaurant for dinner.
Choice B: The new guests coming in the famous steakhouse were four blokes.
For the given two options, choose the one that is more likely to cause the occurrence of the premise.
Use the following JSON format to output your answer:
```json
{
  "correct choice ID": /* one of A, B */
}
```

(*@\textbf{Generated Pseudo-Question:}@*)
Premise: The majority of the passengers on the flight slept during the journey.
Choice A: The flight attendants provided exceptional service and entertainment options.
Choice B: The flight experienced turbulence and delays due to bad weather.

```json
{
  "correct choice ID": "B"
}
```

(*@\textbf{Generated Pseudo-Response:}@*)
The correct choice is B: The flight experienced turbulence and delays due to bad weather. This choice is supported by the premise as it explains why the majority of the passengers slept during the journey likely due to the uncomfortable conditions caused by turbulence and the delay in reaching their destination. Choice A is not supported by the premise and does not explain why the passengers slept during the journey.

\end{lstlisting}\\
\bottomrule
\end{tabular}
\caption{}
\label{cs2}
\end{table*}

\clearpage
\section{Prompts}\label{appendix:prompts}

\captionsetup[table]{name=Prompt} % 重命名
\captionsetup{labelformat=empty} % 不显示Cation但保留ref能力
\setcounter{table}{0} % 重新计数，下一个从1开始。

\begin{table*}[h] \small % prompt
\centering
\begin{tabular}{p{\linewidth}}
\toprule
\textbf{Prompt 1: generate the corresponding task type and task description of the user question.}\\
\midrule\begin{lstlisting}[escapeinside={(*@}{@*)}]
You are an advanced task type induction agent capable of naming a task and describing its goals based on an example of the task.
The description of the task goals should be abstract, general, and essential, avoiding any specifics about how the problem is described or the variable elements within it, as the same task can be described in various ways.
Use the following JSON format to output task name and task descriptions:
```json
{
  "task name": ,
  "task description":
}
```
<Task Example >
(*@\sethlcolor{lightblue}\hl{\textbf{[user question]}}@*)
</Task Example >
\end{lstlisting}\\\bottomrule
\end{tabular}
\caption{}
\label{pt:p1}
\end{table*}

\begin{table*}[b] \small % prompt
\centering
\begin{tabular}{p{\linewidth}}
\toprule
\textbf{Prompt 2: determine whether the target task is identical to one of the candidate tasks in memory.}\\
\midrule\begin{lstlisting}[escapeinside={(*@}{@*)}]
<Target Task>
(*@\sethlcolor{lightblue}\hl{\textbf{[task description of the target task]}}@*)
</Target Task>

<Candidate Task 1>
(*@\sethlcolor{lightblue}\hl{\textbf{[task description of the 1st candidate task]}}@*)
</Candidate Task 1>

<Candidate Task 2>
(*@\sethlcolor{lightblue}\hl{\textbf{[task description of the 2nd candidate task]}}@*)
</Candidate Task 2>

(*@\sethlcolor{lightblue}\hl{\textbf{[...the remaining candidate tasks...]}}@*)

You are an excellent task identifier, capable of determining whether the target task is identical to one of the above candidate tasks.
If no such candidate tasks exist, or if you are unsure, please return -1.
You must carefully avoid selecting any candidate task that are not completely identical to the target task.
Please use the following JSON format to output the selected candidate task:
```json
{
"selected task id": /* -1 or ID of the selected candidate task. */
}
```
\end{lstlisting}\\\bottomrule
\end{tabular}
\caption{}
\label{pt:p2}
\end{table*}

\begin{table*}[b] \small % prompt
\centering
\begin{tabular}{p{\linewidth}}
\toprule
\textbf{Prompt 3: select source tasks for the target task during experience transfer.}\\
\midrule\begin{lstlisting}[escapeinside={(*@}{@*)}]
<Target Task>
(*@\sethlcolor{lightblue}\hl{\textbf{[task description of the target task]}}@*)
</Target Task>

<Candidate Task 1>
(*@\sethlcolor{lightblue}\hl{\textbf{[task description of the 1st candidate task]}}@*)
</Candidate Task 1>
\end{lstlisting}\\
\bottomrule
\end{tabular}
\caption{}
\label{pt:p3}
\end{table*}

\begin{table*}[h] \small % prompt
\centering
\begin{tabular}{p{\linewidth}}
\toprule
\textbf{continued from the above content.}\\
\midrule
\begin{lstlisting}[escapeinside={(*@}{@*)}]
<Candidate Task 2>
(*@\sethlcolor{lightblue}\hl{\textbf{[task description of the 2nd candidate task]}}@*)
</Candidate Task 2>

(*@\sethlcolor{lightblue}\hl{\textbf{[...the remaining candidate tasks...]}}@*)

You are an outstanding source task retriever, capable of discovering source tasks related to the target task from the above candidate tasks.
The experience gained from solving the source tasks should be transferable to the target task.
Use the following JSON format to output the selected source tasks:
```json
{
"selected task ids": [ /* ids of selected source tasks. If there are no suitable source tasks, please return an empty list. */ ]
}
```
\end{lstlisting}\\\bottomrule
\end{tabular}
\caption{}
\end{table*}

\setcounter{table}{3}
\begin{table*}[h] \small % prompt
\centering
\begin{tabular}{p{\linewidth}}
\toprule
\textbf{Prompt 4: transfer the experience of multiple source tasks to the target task.}\\
\midrule\begin{lstlisting}[escapeinside={(*@}{@*)}]
You are an excellent experience transfer agent, adept at transferring experience from one or more source tasks to the target task.
Here is the task description of the target task, as well as the task description and task experience of source tasks.

<Target Task>
(*@\sethlcolor{lightblue}\hl{\textbf{[task description of the target task]}}@*)
</Target Task>

<Source Task 1>
Task Description:
(*@\sethlcolor{lightblue}\hl{\textbf{[task description of the 1st source task]}}@*)
Task Experience:
(*@\sethlcolor{lightblue}\hl{\textbf{[task experience of the 1st source task]}}@*)
</Source Task 1>

<Source Task 2>
Task Description:
(*@\sethlcolor{lightblue}\hl{\textbf{[task description of the 2nd source task]}}@*)
Task Experience:
(*@\sethlcolor{lightblue}\hl{\textbf{[task experience of the 2nd source task]}}@*)
</Source Task 2>

(*@\sethlcolor{lightblue}\hl{\textbf{[...the remaining source tasks...]}}@*)

Please follow the steps below to transfer experience:

Step 1: Task Understanding
Thoroughly understand the target task and source tasks, clearly identifying the commonalities and differences between them.

Step 2: Identify General Experience
Extracting general experience from the source tasks that can also be applied to the target task, especially insights that are common across multiple source tasks.
Avoid using task-specific experience from the source tasks that may not be relevant to the target task.
Be cautious of experience effective in the source tasks but could lead to errors in the target task.
Pay attention to the differences between the source and target tasks.
\end{lstlisting}\\
\bottomrule
\end{tabular}
\caption{}
\label{pt:p4}
\end{table*}

\begin{table*}[h] \small % prompt
\centering
\begin{tabular}{p{\linewidth}}
\toprule
\textbf{continued from the above content.}\\
\midrule
\begin{lstlisting}[escapeinside={(*@}{@*)}]
Step 3: Experience Adaptation
Adapt the general experience identified in Step 2 to the target task, adjusting for aspects that do not align perfectly with the target task's conditions and meeting the specific requirements of the target task.
Ensure that the experience provided are CLEAR, DETAILED, and GENERALLY APPLICABLE to unseen examples in the target task.
Use the following JSON format to output the adapted experience:
```json
{
"How to better accomplish the task or avoid low-quality responses": [ no more than 20 insights ],
"The specific process for handling this task": [ no more than 20 insights ]
}
```

Let's think step by step.
\end{lstlisting}\\\bottomrule
\end{tabular}
\caption{}

\end{table*}

\setcounter{table}{4}
\begin{table*}[h] \small % prompt
\centering
\begin{tabular}{p{\linewidth}}
\toprule
\textbf{Prompt 5: combine and deduplicate two sets of experience for the same task.}\\
\midrule\begin{lstlisting}[escapeinside={(*@}{@*)}]
<Target Task>
(*@\sethlcolor{lightblue}\hl{\textbf{[task description of the target task]}}@*)
</Target Task>

<Existing Experience>
{
"How to better accomplish the task or avoid low-quality responses":
(*@\sethlcolor{lightblue}\hl{\textbf{[list all the unordered suggestions from two sets of experience.]}}@*),
"Task Processing Flow 1": (*@\sethlcolor{lightblue}\hl{\textbf{[the ordered procedure from the first set of experience.]}}@*),
"Task Processing Flow 2": (*@\sethlcolor{lightblue}\hl{\textbf{[the ordered procedure from the second set of experience.]}}@*)
</Existing Experience>

You are an excellent experience refiner. Please help me refine the above existing experience related to the target task.
1. For "How to better accomplish the task or avoid low-quality responses", please integrate insights by combining those that are closely related and eliminating any repetitions.
2. Please integrate the above "Task Processing Flow 1" and "Task Processing Flow 2" into one unified workflow process. Ensure that the primary goals and functionality of both original processes are preserved; Effectively resolve possible conflicts or overlaps between the two processes.
Use the following JSON format to output refined target task experience:
```json
{
"How to better accomplish the task or avoid low-quality responses": [ no more than 20 insights ],
"The specific process for handling this task": [ no more than 20 insights ]
}
```
\end{lstlisting}\\\bottomrule
\end{tabular}
\caption{}
\label{pt:p5}
\end{table*}

\begin{table*}[h] \small % prompt
\centering
\begin{tabular}{p{\linewidth}}
\toprule
\textbf{Prompt 6: generate a new question of the target task type based on the reference web text.}\\
\midrule\begin{lstlisting}[escapeinside={(*@}{@*)}]
<Reference Text>
(*@\sethlcolor{lightblue}\hl{\textbf{[reference text retrieved from the internet]}}@*)
</Reference Text>
\end{lstlisting}\\
\bottomrule
\end{tabular}
\caption{}
\label{pt:p6}
\end{table*}

\begin{table*}[h] \small % prompt
\centering
\begin{tabular}{p{\linewidth}}
\toprule
\textbf{continued from the above content.}\\
\midrule
\begin{lstlisting}[escapeinside={(*@}{@*)}]
<Example Question>
(*@\sethlcolor{lightblue}\hl{\textbf{[The example question of the target task, i.e., the input user question of our framework]}}@*)
</Example Question>

<Task Type of the Example Question>
(*@\sethlcolor{lightblue}\hl{\textbf{[task description of the target task]}}@*)
</Task Type of the Example Question>

You are an excellent questioner.
Please carefully read the reference text provided above and formulate a new question based on it.
The new question must maintain the same expression style, structure, and required output format as the example question.
The new question must belong to the same task type of the example question.
The new question must be well-defined, with a complete and clear description that can be answered and at least one correct answer exists.
You are forbidden from providing answers to your new question.
Use the following format to output your answer:
<New Question>
/* Your new question. */
</New Question>
\end{lstlisting}\\\bottomrule
\end{tabular}
\caption{}
% \label{pt:p6}
\end{table*}

\setcounter{table}{6}
\begin{table*}[h] \small % prompt
\centering
\begin{tabular}{p{0.98\linewidth}}
\toprule
\textbf{Prompt 7: during the autonomous practic process, generate a thought process and answer to the generated new question based on experience.}\\
\midrule\begin{lstlisting}[escapeinside={(*@}{@*)}]
<Task Experience>
(*@\sethlcolor{lightblue}\hl{\textbf{[experience of the target task]}}@*)
</Task Experience>
Please refer to the above experience to answer the following question.
(*@\sethlcolor{yellow}\hl{\textbf{\# The above part is omitted when the experience is empty.}}@*)

(*@\sethlcolor{lightblue}\hl{\textbf{[a generated new question]}}@*)

Please provide specific, detailed, and comprehensive steps of your thought.
\end{lstlisting}\\\bottomrule
\end{tabular}
\caption{}
\label{pt:p7}
\end{table*}

\begin{table*}[h] \small % prompt
\centering
\begin{tabular}{p{\linewidth}}
\toprule
\textbf{Prompt 8: based on the reference text, check if the response to the question is correct.}\\
\midrule\begin{lstlisting}[escapeinside={(*@}{@*)}]
<Reference Text>
(*@\sethlcolor{lightblue}\hl{\textbf{[reference text retrieved from the internet]}}@*)
</Reference Text>

<Target Question>
(*@\sethlcolor{lightblue}\hl{\textbf{[the generated new question]}}@*)
</Target Question>

<Reasoning Process and Answer>
(*@\sethlcolor{lightblue}\hl{\textbf{[the thought process and answer of the new question]}}@*)
</Reasoning Process and Answer>
\end{lstlisting}\\
\bottomrule
\end{tabular}
\caption{}
\label{pt:p8}
\end{table*}

\begin{table*}[h] \small % prompt
\centering
\begin{tabular}{p{\linewidth}}
\toprule
\textbf{continued from the above content.}\\
\midrule
\begin{lstlisting}[escapeinside={(*@}{@*)}]
You are an outstanding checker, skilled at examining the reasoning process and the correctness of the answer of the target question based on the reference text.
Pay close attention to whether the reasoning process and the answer are consistent or inconsistent with the reference text.
Use the following JSON format to output your opinion:
```json
{
"correctness": /* "correct", "wrong" or "inconclusive" */
}
```

Let's think step by step.
\end{lstlisting}\\\bottomrule
\end{tabular}
\caption{}
% \label{pt:p8}
\end{table*}

\setcounter{table}{8}
\begin{table*}[h] \small % prompt
\centering
\begin{tabular}{p{\linewidth}}
\toprule
\textbf{Prompt 9: summarize the task-solving experience from examples with correct or incorrect answers.}\\
\midrule\begin{lstlisting}[escapeinside={(*@}{@*)}]
You are an excellent experiential summarizer, adept at extracting task-solving insights from examples of the target task.
Here are several target task examples with correct or incorrect answers:
<Correct Example 1>
<Question>
(*@\sethlcolor{lightblue}\hl{\textbf{[the generated new question]}}@*)
</Question>
<Reasoning Process and Answer>
(*@\sethlcolor{lightblue}\hl{\textbf{[the thought process and answer of the new question]}}@*)
</Reasoning Process and Answer>
</Correct Example 1>

(*@\sethlcolor{lightblue}\hl{\textbf{[...the remaining correct examples...]}}@*)

<Incorrect Example 1>
<Question>
(*@\sethlcolor{lightblue}\hl{\textbf{[the generated new question]}}@*)
</Question>
<Reasoning Process and Answer>
(*@\sethlcolor{lightblue}\hl{\textbf{[the thought process and answer of the new question]}}@*)
</Reasoning Process and Answer>
</Incorrect Example 1>

(*@\sethlcolor{lightblue}\hl{\textbf{[...the remaining incorrect examples...]}}@*)

Based on the examples provided above, please follow the steps below to summarize the experience:

Step1: Observe and Analyze the Examples
Summarize the commonalities in the correct examples, identify patterns in the incorrect examples, and compare the differences between the correct and incorrect examples.

Step2: Summarize Experience
Based on the observations and analysis from the Step1, summarize task-solving insights.
Ensure that the insights provided are CLEAR, DETAILED, and are GENERALLY APPLICABLE to unseen examples of the target task.
Use the following JSON format to output the summarized experience:\end{lstlisting}\\
\bottomrule
\end{tabular}
\caption{}
\label{pt:p9}
\end{table*}

\begin{table*}[h] \small % prompt
\centering
\begin{tabular}{p{\linewidth}}
\toprule
\textbf{continued from the above content.}\\
\midrule
\begin{lstlisting}[escapeinside={(*@}{@*)}]
```json
{
"How to better accomplish the task or avoid low-quality responses": [ no more than 20 insights ],
"The specific process for handling this task": [ no more than 20 insights ]
}
```

Let's think step by step.
\end{lstlisting}\\\bottomrule
\end{tabular}
\caption{}

\end{table*}

\setcounter{table}{9}
\begin{table*}[h] \small % prompt
\centering
\begin{tabular}{p{\linewidth}}
\toprule
\textbf{Prompt 10: think the question based on experience and respond to the user.}\\
\midrule\begin{lstlisting}[escapeinside={(*@}{@*)}]
<Experience>
[How to better accomplish the task or avoid low-quality responses]:
(*@\sethlcolor{lightblue}\hl{\textbf{[list the unordered suggestions from the experience.]}}@*)
[The specific process for handling this task]:
(*@\sethlcolor{lightblue}\hl{\textbf{[list the ordered procedure from the experience.]}}@*)
</Experience>
Please refer to the above experience to answer the following question.

(*@\sethlcolor{lightblue}\hl{\textbf{[the input user question of our framework]}}@*)
\end{lstlisting}\\\bottomrule
\end{tabular}
\caption{}
\label{pt:p10}
\end{table*}

\begin{table*}[h] \small % prompt
\centering
\begin{tabular}{p{\linewidth}}
\toprule
\textbf{Prompt 11: directly generate task-solving experience for the input question.}\\
\midrule\begin{lstlisting}[escapeinside={(*@}{@*)}]
You are an excellent advisor, skilled in providing task-solving insights for the target task.
<Target Task>
(*@\sethlcolor{lightblue}\hl{\textbf{[the input question]}}@*)
</Target Task>

Please give your suggestions.
Ensure that the insights provided are CLEAR, DETAILED.
Use the following JSON format to output:
```json
{
"How to better accomplish the task or avoid low-quality responses": [ your insights ],
"The specific process for handling this task": [ your insights ]
}
```
\end{lstlisting}\\\bottomrule
\end{tabular}
\caption{}
\label{pt:p11}
\end{table*}

\clearpage
\section{Examples of Our Framework}\label{appendix:case_study_our}
In this section, we demonstrate examples of our framework.
Due to the memory limits on arXiv's compilation, subsequent pages are available through the following link: \url{https://drive.google.com/file/d/17zc4oUuvq2-BsaZ55Zd9O13TCVU49RxF/view?usp=share_link}

\end{document}